# FINGER BIOMETRIC RECOGNITION WITH FEATURE SELECTION

ASISH BERA [1], DEBOTOSH BHATTACHARJEE [2], AND MITA NASIPURI [2]

asish.bera@gmail.com, debotosh@ieee.org, and mitanasipuri@gmail.com

[1] Department of Computer Science and Engineering, Haldia Institute of Technology, Haldia-721657.

[2] Department of Computer Science and Engineering, Jadavpur University, Kolkata-700032.

**Abstract**

Biometrics is indispensable in this modern digital era for secure automated human authentication in various fields of machine learning and pattern recognition. Hand geometry is a promising physiological biometric trait with ample deployed application areas for identity verification. Due to the intricate anatomic foundation of the thumb and substantial inter-finger posture variation, satisfactory performances cannot be achieved while the thumb is included in the contact-free environment. To overcome the hindrances associated with the thumb, four finger-based (excluding the thumb) biometric approaches have been devised. In this chapter, a four-finger based biometric method has been presented. Again, selection of salient features is essential to reduce the feature dimensionality by eliminating the insignificant features. Weights are assigned according to the discriminative efficiency of the features to emphasize on the essential features. Two different strategies namely, the global and local feature selection methods are adopted based on the adaptive forward-selection and backward- elimination (FoBa) algorithm. The identification performances are evaluated using the weighted *k*-nearest neighbor (*wk*-NN) and random forest (RF) classifiers. The experiments are conducted using the selected feature subsets over the 300 subjects of the Bosphorus hand database. The best identification accuracy of 98.67%, and equal error rate (EER) of 4.6% have been achieved using the subset of 25 features which are selected by the rank-based local FoBa algorithm.

## 1 INTRODUCTION

Biometrics is inevitable in diverse application areas of pattern recognition and machine learning since the last five decades [1]. Its inherent benefits surmount its applicability in secure human authentication than the common password- and token-based verification methods. Several obtrusive physiological (e.g., fingerprint, face, hand geometry, hand vein, retina, etc.) and behavioral (e.g., gait, keystroke, signature, etc.) modalities have been explored for finding a reliable biometric solution [2]. Several biometric traits are deployed in the government (national ID card), financial (banking transaction), and industrial (attendance maintenance) purposes for secure authentication. Moreover, biometrics is a viable tool for individualization in the real-time systems, surveillance, forensic investigation, internet of things (IoT), and other online-based services. However, a single trait is not always adequate to render satisfactory accuracy with a more significant population especially, thousands, or even more. In such cases, multi-modal biometrics is an alternative solution [3]. Several multibiometric and fusion-based approaches have been implemented for performance and security enhancement. Astoundingly, billions of devices have been connected through the internet of things (IoT) with the biometric verification facilities [4]. It has been predicted that more than 20 billions of devices will be connected through the IoT in 2020 [5]. Thus, biometric is essentially an invulnerable tool for human authentication in this contemporary society [6].

Personal verification using the hand biometric properties also renders a robust solution which was pioneered with the two earliest US patents in the 1970s [7-8]. Then, after almost three decades, prospects of hand geometry have been investigated rapidly, with the state-of-the-art method proposed by *Sanchez et al.* in 2000 [9]. The image acquisition framework of that system is based on the guiding pegs. After the hand image preprocessing, geometric measurements of the hand are computed, and salient features are selected for experiments. Later on, the hand biometric approaches have been presented with the hand images acquired in a contact-free manner, i.e., without the assistance of any rigid peg. According to the study, it can be stated that most of the unconstrained hand biometric approaches have been developed since after the year 2000. A compendious study was summarized by *Duta* in 2009 [10]. Most of the novel methods have been emphasized on the pose-flexibility in a contactless environment. Moreover, several unconstrained and contact-free proprietary hand databases have been created in different nations to facilitate the researchers for devising novel techniques. In this regard, other than the 2-D hand images [11-12], databases of the 3-D [13], infrared [14], hyperspectral [15], and synthetic [16] hand images have been tested.

Every trait carries its pros and cons which signifies the specificity in multiple real-world applications. Hand biometrics also bears specific intrinsic characteristics, and provides non-intrusive identity verification solutions reliably. The principal advantages and drawbacks of hand biometrics are summarized in Table 1. However, regardless of several limitations, the usefulness of this trait is attractive; and deployed ubiquitously for automatic attendance management in the nuclear plant, identity verification for international border control, and others. Due to the inherent limits, this mode is mainly suitable for verification purposes rather than identification of a larger population. The fundamental weakness pertinent to the posture variation cannot be eradicated accurately because of the intricate anatomic foundation of the thumb finger. Hence, the difficulty is avoided just by removing the thumb from feature computation [17-18]. In some approaches, only these four fingers are segmented while the thumb and palm are neglected before feature extraction. This modality is known as the four-finger biometrics that also provides equivalent or even better accuracies compared to the five-finger based methods.

Table 1: The pros and cons of hand biometrics

| Advantages | Disadvantages |
|---|---|
| a) Low-cost of the sensors, such as a traditional scanner or camera.<br>b) Size of feature template is small, typically only a few bytes.<br>c) User-friendliness due to simple operations for verification.<br>d) High user acceptability and less invasiveness. | a) Interface contact between the hand and device may raise hygienic issues.<br>b) Intra-class posture variation is significantly more than inter-class.<br>c) Low uniqueness and universality for identification.<br>d) Larger device surface prevents its deployment in smaller hand-held devices, like Smartphone. |

A traditional biometric system is divided into four key modules, namely, the image acquisition (or sensor) module, feature extraction (and selection) module, template matching module, and decision-making module. Earlier than feature computation, a robust preprocessing method is followed for finger segmentation. Next, either the geometric measurements or the shape-based descriptors from the fingers are computed [6, 19]. The pixel-level information is transformed into a descriptor, such as the Fourier (FDs) [17], shape-contexts (SC) [18], wavelet (WDs) [20], scale invariant feature transform (SIFT) [21], and others. Both categories of the features are equally imperative to represent an identity uniquely. Notably, a conflation of both types of features can also endow satisfactory results. Now, all the defined features (i.e., either geometric or shape-based) are not capable enough to contribute acceptably in performance evaluation. Hence, only the significant features should be chosen using a suitable feature selection technique in the training phase [22-23]. The aim is not only to compute a diverse set of salient features, but, also to select a minimal subset of discriminative features. The insignificant features are discarded which implies the dimensionality reduction of the input feature set. A pictorial representation of the finger biometric system with the precise tasks during the training and testing phases is shown in Fig.1.

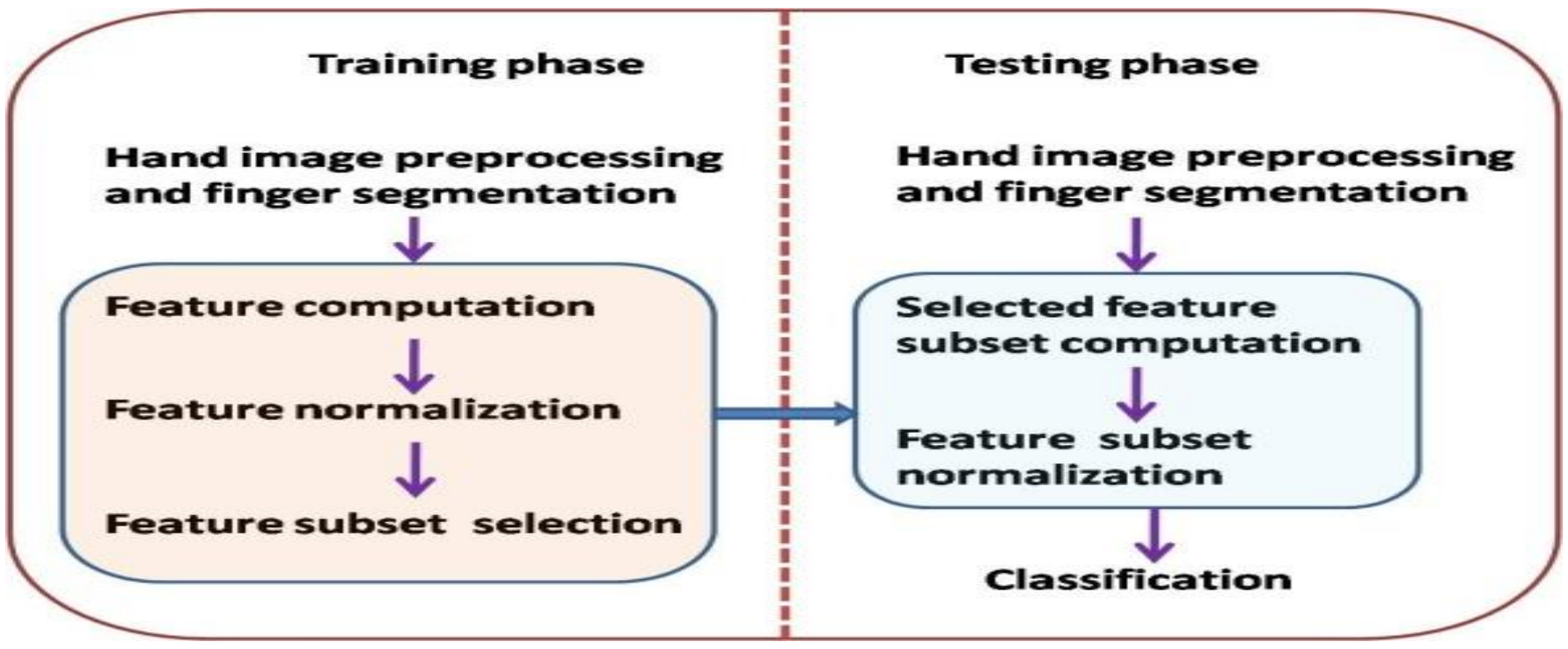

Figure 1: A finger biometric approach with feature selection strategy

In the context of hand biometrics, different schemes for feature selection have been implemented. The principal objectives are discussed next.

## 1.1 MOTIVATION

It can be averred that other than a simple unimodal hand biometric approach, mainly, three different types of schemes are adopted in literature for improving the robustness and accuracy of a hand biometric system. Explicitly, it can be categorized as

a) *multimodal scheme:* two or more modalities are employed, such as the hand and face [24]. The feature extraction methods differ according to the modalities applied. The matching scores of different systems or traits are combined at the score-level or decision-level fusion.
b) *mono-modal with a fusion strategy:* a suitable fusion technique is applied at any stage of implementation, e.g., matching score-level [17, 47]. A diverse set of feature extraction method on the underlying trait has also been emphasized.
c) *unimodal with a feature selection technique:* only relevant features (e.g., hand geometric) are selected, and extraneous features are discarded to reduce feature dimensionality using any apposite metric, such as the mutual information (MI), correlation, etc. [19, 48].

These schemes have been experimented over various feature sets of fingers or hand-shape in different existing works. The finger biometric features emphasize on the local characteristics of fingers. It has been stated earlier that a sizeable intra-class posture variation of the thumb affects the accuracy while included during experimentation [25]. Herein, a finger geometry recognition approach with the feature selection algorithms is described. The principal objectives of this chapter are summarized as follows.

i) An exhaustive vignette on the state-of-the-art methods on finger biometric authentication with an emphasis on the feature selection techniques is presented.

ii) Finger geometric feature extraction and selection of global and local relevant features that optimize the feature space, and improve the performance is described.

iii) An in-depth experimental analysis of the identification and verification performance evaluation with the weighted and selected feature subspaces defined over the Bosphorus database is illustrated.

The rest of this chapter is organized as follows. Section 2 represents a study on the related hand and finger state-of-the-art biometric approaches. Preprocessing methodology and feature computation of finger biometric modality is described in Section 3. Feature selection algorithms are described in Section 4. The experimental conception and results are analyzed in Section 5. Lastly, a final remark is drawn in Section 6.

## 2 RELATED WORKS

Hand biometrics has been excogitated in several possible research directions. Based on the finger biometric feature extraction module, existing approaches can be classified into four categories, pictorially shown in Fig.2. A summary of those methods is briefed here.

*Feature resource:* the raw hand images are collected using a scanner or a camera from the users at different imaging sessions. After the preprocessing, a uniform representation is followed for all the images, and the hand is segmented. This process is also known as hand normalization [11, 46]. Next, a set of specific features are computed from the whole hand and/or particular to the fingers. In the case of finger biometry, the fingers are segmented and separated from the actual hand. Next, the features are computed from the normalized fingers. It can be noted that the anatomic intricacy of the thumb precludes its feature computation in some cases that implicate the significance of the four-finger biometrics.

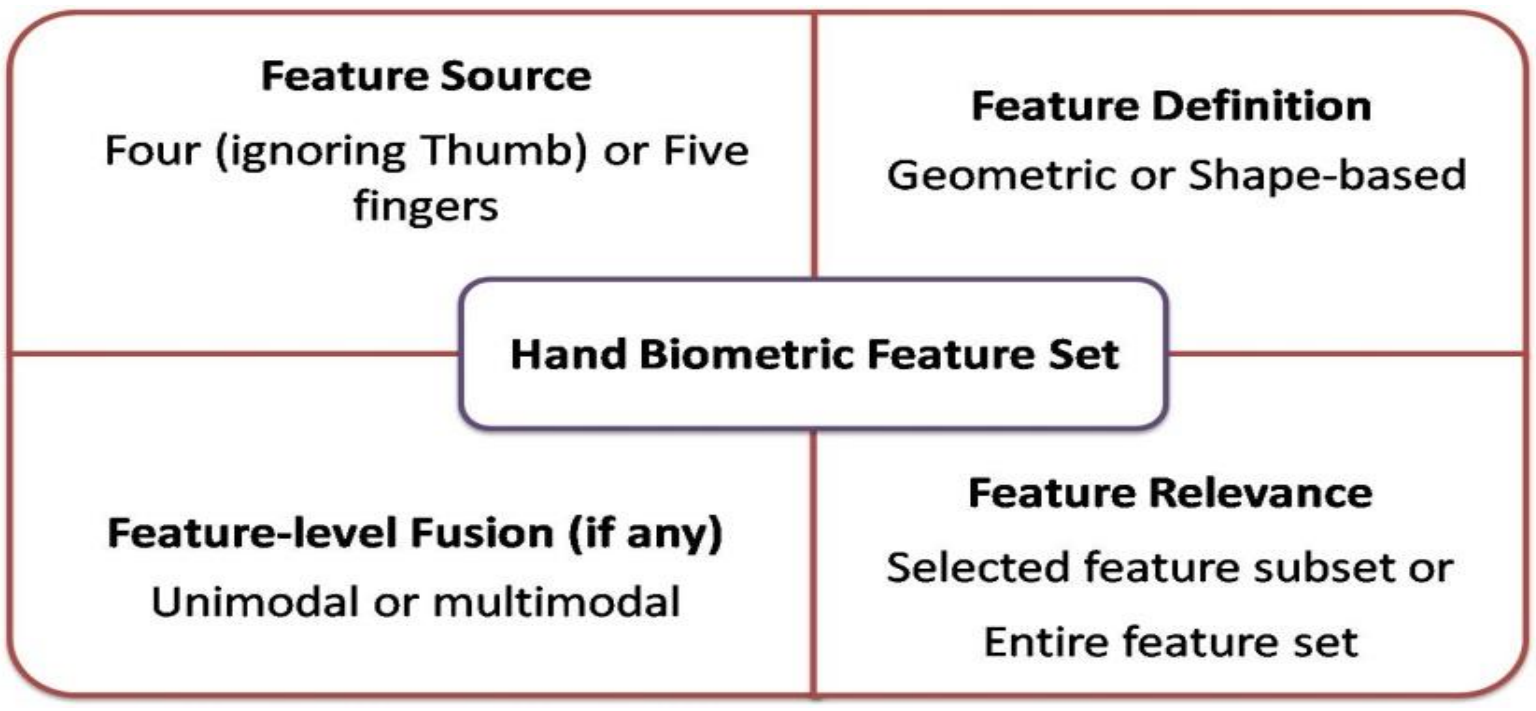

Figure 2: Different approaches on hand biometric feature set

*Feature definition:* a feature set is computed based on the geometric measurements (e.g., area, length, width, etc.) at various aspects and/or shape-descriptors using different spatial transformation techniques such as the Fourier transform (FDs), wavelet transform (WDs), and others [6, 19]. The dimension of a geometric feature set (usually, within hundred) is somewhat smaller than shape-based features (generally, few hundred or more). Thus, dimensionality reduction method (such as principal component analysis) is usually followed, and only the essential attributes are chosen while the rest are discarded. Notably, geometric feature extraction and matching tasks are slightly more straightforward than the silhouette-based features due to the smaller feature space regarding the computational complexity. The main advantage of a shape-descriptor is the invariance property under one or more affine transformation(s) such as the translation, rotation, scaling, and others [21]. Some common feature characteristics are summarized in Table 2.

Table 2: Finger biometric feature definition

## Geometric Features [19, 25]:

*Area (A):* the number of boundary pixels of the hand (or finger).

*Length:* the Euclidean distance between the finger-tip pixel and mid-pixel of the finger-baseline.

*Widths:* the breadths of a finger at different latitude along its length.

*Extent:* the ratio of the area and minimum bounding rectangle area (MBRA), given as *A/MBRA*

*Perimeter (P):* the cumulative distance between each pair of adjacent contour pixels.

*Major-axis length and minor-axis length*: the major-axis and minor-axis lengths of the fitted ellipse over every finger. The biaxial lengths are invariant to rotation and translation.

*Solidity:* the ratio of area and convex hull area(CA), given as *A/CA*

*Circularity:* defined as $4\pi A/P^2$

*Compactness*: measured as $P^2/A$

*Inter-finger angle:* defined concerning the pivotal finger axes or located vital points.

*Equivalent diameter*: computed as $(4\times Area/\pi)^{0.5}$

*Distance from Centroid(DC):* the distance from the centroid ($x_c$, $y_c$) of a finger to a contour pixel (x,y), defined as $[(x - x_c)^2 + (y - y_c)^2]^{0.5}$

## Shape-based Features:

*Fourier descriptor (FD):* the magnitudes of the points after the discrete Fourier transformation [17].

*Wavelet descriptor (WD):* the coefficients after the wavelet transformation with multi-resolution analysis [20].

*Shape context (SC):* defined with *n* pixels representing a shape. For any pixel, a histogram of the relative coordinates of remaining *(n-1)* pixels is computed. The histogram describes the SC of that pixel, and the bins of the histogram are defined in a log-polar space [18].

*Scale-invariant Feature Transform (SIFT):* the local maxima and minima key points using Gaussian linear transformation at various scale-spaces. Noisy and contour pixels are removed [21].

*Hu moment invariant (HMI):* defined with the second and third normalized central moments; consisting of six complete orthogonal invariants and one skew orthogonal invariant based on algebraic invariants, and representing the statistical properties [19].

*Zernike moment invariant(ZMI):* based on a set of complex polynomials that form a complete orthogonal set over the interior of a unit circle, and defined as the projection of the image on these orthogonal basis functions [26].

*Angular radial transform (ART):* based on the polar coordinate system in which the sinusoidal basis functions are defined on a unit disc [27].

| Invariance property | DC | FD | WD | SC | HMI | ZMI | ART | SIFT |
|---|---|---|---|---|---|---|---|---|
| Translation | √ | √ | √ | √ | √ | √ | √ | √ |
| Rotation | | √ | √ | √ | √ | √ | √ | √ |
| Scaling | | √ | √ | √ | √ | √ | √ | √ |
| Noise | | | | | | | √ | √ |
| Illumination | | | | | | | | √ |
| Distortion | | | | | | | | √ |

Table 3: Hand biometric approaches with feature selection

| Sl. | Author | Feature extraction | Feature selection method | Experimentation |
|---|---|---|---|---|
| 1 | R.S. Reillo *et al.*, 2000, [9] | 31 geometric features are measured from hand. It includes the widths and heights of fingers, deviations, and inter-finger angles. | *Class variability ratio:* relevance of the features is estimated by the proportion of the inter-class variability and intra-class variability. The class-variability is determined using the standard deviation and the mean of feature and number of classes. A higher value ratio implies a more relevant feature. | Database: 20 persons. 25 discriminative features are selected. 97% classification accuracy and much lesser than 10% EERs have been achieved using Gaussian mixture model (GMM). |
| 2 | Kumar *et al.*, 2005, [28] | 23 geometric features are computed from hand. It includes the perimeter, solidity, extent, palm length and width, finger length and widths, etc. | *Correlation-based feature selection (CFS):* it is a classifier-independent algorithm, and the *Pearson* correlation metric is used. The CFS has effectively reduced the number of features with maintaining the similar performances. | Database: 100 subjects. 15 relevant features are selected. Identification accuracy of 87.8% has been achieved using the logistic model tree (LMT) classifier. |
| 3 | Baena *et al.*, 2013, [19] | Total 403 features are computed from each finger and hand. It defines the finger length, compactness, widths, rectangularity, extent, solidity, FDs, HMI, and others. | *Genetic algorithm (GA) with Mutual information (GA-MI):* GA is applied for feature selection and MI is employed in the fitness function to find out the correlation between a pair of features and to eliminate the redundancy among features. 100 different executions of GA have been tested for relevant feature selection. | Databases: GPDS (144 subj.), IITD (137 subj.), and CASIA (100 subj.). Selects about 50 features, identification precision about 97% has been achieved with the latter two databases using GA-LDA and EERs of 4-5%. |
| 4 | Bera *et al.*, 2017, [25] | 30 geometric features per finger are defined. It includes the area, solidity, extent, major and minor axes lengths of an ellipse fitted over the finger, 10 CDs, widths at 10 different positions of the finger, etc. | *Rank-based forward-backward (RFoBa):* A rank is assigned to every feature according to its relevance. Based on the rank, the forward selection is followed at first. Then, the backward elimination is applied successively to formulate two optimal subsets of selected features. | Database: Bosphorus (638, subjects). With the four fingers and 12 selected features per finger of the right-hand, identification success of 96.56% using the random forest classifier and EER of 7.8% have been achieved. |

*Feature relevance:* the uniqueness of a feature determines its significance in the feature set. The relevant features are selected to define the identity while the insignificant and noisy features are discarded. Feature relevance exhibits the aptness for dimensionality reduction too. The relevant features are chosen, and an optimal subset is formulated using a suitable metric, such as the MI. The number of useful features is a general condition for subset formulation and stopping criterion. The relevance computation is explained in Section 4. Few approaches on hand and finger biometrics have the benefits of feature selection are summarized in Table 3.

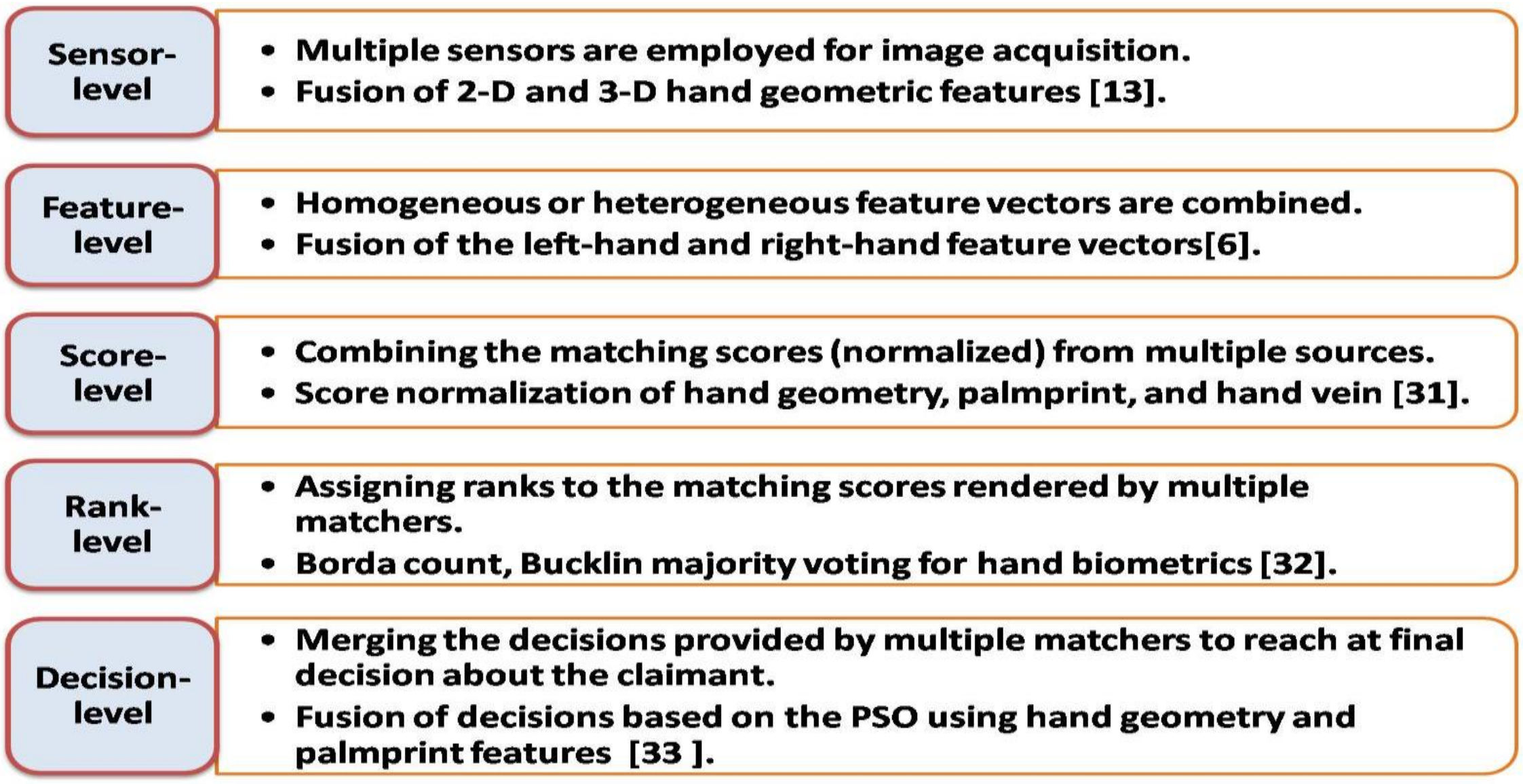

Figure 2: Different implementation levels of fusion-based hand biometric approaches

*Feature-level fusion:* another paradigm is to combine different feature spaces. Homogeneous feature sets (e.g., two different hand geometric feature vectors of the same subject) are conflated with the mean values of corresponding features. Compatible feature sets are more comfortable to manipulate, and to formulate a uniform feature template [3]. Alternatively, heterogeneous feature templates (e.g., fingerprint minutiae and hand geometric features) are concatenated after normalizing them into a standard feature space. Combining two different vectors into a single one also increases the overall dimension of the feature vector. Another solution is to match the heterogeneous feature vectors separately, and fuse the matching scores to render the final score or to take a decision about the claimed identity [29-31]. A vignette about the various levels of fusion techniques in the context of hand biometrics is illustrated in Fig.2.

In conjunction with the hand and finger geometry, some other hand-based modalities are fused at various level of design, such as the fingerprint, palmprint, hand vein, finger knuckle print, and others [34]. Also, a non-hand based trait, like the facial pattern is also fused [24].

Furthermore, the other modules (such as the sensor module) have also been improved for a robust solution regarding the accuracy and computational time. However, it can be averred that the approaches directed toward feature selection techniques need more attention than traditional all feature usability approach. Hence, this chapter is also concerned about this objective. In the next section, the preprocessing of an input hand image is described.

## 3 Preprocessing methodology of finger geometry

The preprocessing of hand images necessitates few successive steps to segment the hand from a darker background. In literature, different techniques have been devised for hand segmentation. For example, the K-means clustering algorithm has been used for hand segmentation in [17]. After segmenting the hand region of interest (ROI), the fingers are isolated and separated from each other in the case of finger biometry. Herein, a study on finger segmentation method from [25, 35] is described briefly.

### 3.1 Elementary preprocessing of hand image

The preprocessing tasks described in different available works follow similitude operations. Firstly, a color hand image ($L_C$) is converted into an equivalent grayscale image ($L_G$). Noise at the sensor-level may be introduced due to environmental factors such as dirt, inconsistent lighting condition, etc. The noise associated with an input image is removed by a suitable filter such as a median or Gaussian filter. Next, the grayscale image is converted into a binary image ($L_B$) using the *Otsu's* method for thresholding [36]. A modification of *Otsu's* approach has also been applied in [17]. The hand contour is obtained from the binary image using an edge detection algorithm, such as the *Sobel's* [19] or *Canny's* edge detector [37]. The hand contour may be composed of several components because of any worn hand accessory such as a wristwatch, bracelet, ring(s), etc. A ring artifact removal method has been followed to overcome the occlusion due to the hand accessories in [11]. Morphological opening corrections are inevitable to recuperate the left out contour pixels and to smooth the silhouette as a connected component of interest. Next, the essential contour component is selected while other insignificant parts (if any) are neglected for further operations.

In a contact-free image acquisition system, a consistent orientation of every hand silhouette is necessary for hand normalization. For rotating the hand contour, ellipse fitting method is followed in some earlier works such as in [20]. To achieve a uniform orientation, an ellipse is fitted around the ROI so that the major-axis passes through the centroid of ROI and coincides with the Cartesian Y-axis. The centroid ($x_c$, $y_c$) is defined based on the moments, $m_{i,j}$, of the binary ROI.

$$(x_c, y_c) = (m_{1,0}/m_{0,0}\ ,\ m_{0,1}/m_{0,0}) \tag{1}$$

The angle of rotation ($\theta$) is estimated as

$$\theta = 0.5 \times \tan^{-1}\left(\frac{2m_{1,1}}{m_{2,0} - m_{0,2}}\right) \tag{2}$$

After rotation of the image containing ROI, the resulting image is denoted as $L_R$. These simple processing steps are common in several state-of-the-art techniques of hand biometrics. However, the finger isolation can be achieved by different methods, briefed next.

An approach for finger separation from the hand ROI is followed in [25, 35]. In that approach, only simple arithmetic and Boolean operations are followed for finger segmentation while some other methods used the reference points for finger ROI extraction. The critical steps of this method are described next, and ideated in Fig.3. Image $L_G$ is rotated with the equal θ degree, and resulting in image $L_\theta$. Images $L_\theta$ and $L_R$ are used for the next operations.

### 3.2 Finger segmentation

Segmentation of the fingers from the entire hand shape is another crucial task. The tip-valley points are located for this purpose. In [17], the finger-valley points and necessary reference points are used for segmenting the four fingers. The radial distance map of the hand contour is a familiar method for localization of the tip-valley points [11, 17]. Contrarily, the localization of reference points is not required for finger isolation in [25]. In this approach, the primary hand contour component is disunited into the left- and right-profile at finger-level. Consecutive steps of operations have been followed to isolate the finger profiles from the entire hand contour component.

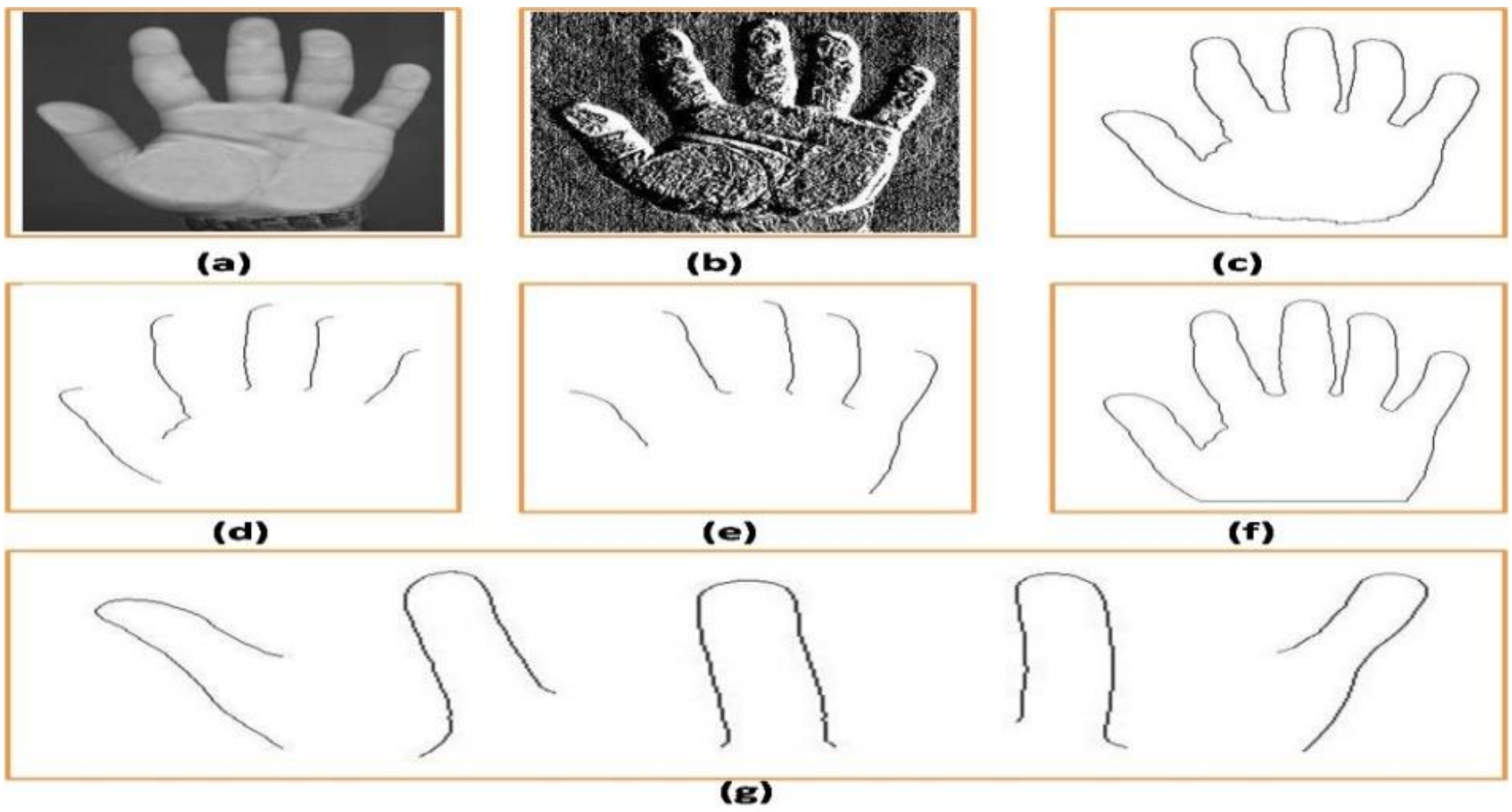

Figure 3: significant outcomes of the preprocessing steps. (a) grayscale hand image, (b) grayscale image transformation, (c) most vital component of the hand contour, (d) left sides of the finger contour profiles, (e) right sides of the finger contour profiles, (f) wrist removal and normalized hand shape, (g) segmented fingers from the left to right direction: thumb, index, middle, ring, and little finger.

In the first step, grayscale intensity variation between every pair of neighbor pixels of $L_\theta$ is computed by subtraction at the pixel-level that follows a similitude operation of computing the first order horizontal gradient magnitude.

$$L_H = \eta\times[\ L_\theta (p, q) - L_\theta (p, q+1)\ ] \qquad \eta=1/2^h\ ;\ h=1,2,3... \tag{3}$$

where, p defines the altitude-index, and q is the breadth-index of $L_\theta$. Next, $L_H$ is converted into a binary image ($L_{BW}$) according to the intensity-profile of pixels in $L_H$. The variations of $L_H$ formation at a different scale for determining the constant value (here, h=1, i.e., η=0.5 in eq.3) have been explained in [35]. The non-zero intensities of $L_H$ are converted into the white (1-value) pixels while the remnant pixels are regarded as the black (0-value). Image $L_H$ contains white-pixels mainly at the right-side of each finger due to the intensity changeover from fore-ground to background.

In the next step, binary image subtraction is applied between $L_R$ and $L_{BW}$ that results in the left-profile (LP) of every finger shape.

$$L_{LP} = L_R - L_{BW} \tag{4}$$

Binary image $L_{LP}$ represents the isolated contour segments at the left-side of all fingers. Morphological corrections are followed such as closing and bridging with a suitable structure element of 3×3 to eliminate an undesirable affect due to noise or intensity alternation.

In the third step, Boolean *xor* operation is applied to retrieve the remnant contour segments at the other side, named as the right-profiles (RP) of the fingers.

$$L_{RP} = L_R \oplus L_{LP} \tag{5}$$

Image $L_{LP}$ contains the minor part of profile segment of the little finger, and $L_{RP}$ comprises the lower portion of contour fragment of the thumb. In the last step, these two lower regions of the respective profile segments are discarded for uniformity in shape representation by using a straight reference line which can be determined in various ways, such as a certain distance from the centroid of the hand. This wrist removal is significant to avert the errors during feature computation [11]. The conflation of the corresponding left- and right-profiles represents a single and segregated finger.

The entire process from an input image to finger isolation is known as finger normalization. In four-finger based approach, after finger segregation, the thumb is excluded explicitly before feature extraction.

### 3.3 Geometric feature set computation

Computation of salient features is another crucial task which may also be inevitably united with some noisy data. The noisy features degrade the performance during testing on the actual feature space. In hand biometrics, the geometric measurements and shape-based descriptors are extracted as features. Herein, a set of only thirteen easily computable geometric features are extracted from each of the normalized four fingers (the number of the feature is mentioned within braces). The feature set includes:

- Area: (1),
- Solidity: (1),
- Equivalent diameter: (1),
- Major-axis and minor-axis lengths of the ellipse fitted on the finger: (2),
- Width per phalanx of a finger: (3), and
- Distances from the centroid of the finger to five equidistant contour pixels: (5).

These geometric features are defined earlier in Table 2. Now, feature normalization is another issue when the magnitudes of the features are in various scalar ranges. It causes a feature $fe_i$ with a higher value can lead over another $fe_j$ with a lower value. It may also be the reason for other features to turn into irrelevant during the experiment. So, all the features are normalized to [0, 1] using the *min-max* rule

$$f_{i,j} = \left(fe_{i,j} - \min(fe_i)\right) / \left(\max(fe_i) - \min(fe_i)\right) \tag{6}$$

where, the $fe_{i,j}$ represents the $i^{th}$ feature of the $j^{th}$ sample, *max(fe*$_i$*)* and *min(fe*$_i$*)* represent the maximum and minimum values of the $i^{th}$ feature, respectively. The normalized features are denoted as $f_{i,j}$ or simply $f_i$ in the set $\mathcal{F} = \{f_i\}_{i=1}^{13}$ which are used for subset selection next.

## 4 Feature selection algorithms

The essential features are selected mainly using the filter and wrapper methods [22, 38]. The embedded and hybrid are two other derived methods. Filter method is classifier independent, because it selects the features irrespective of any classification model. Mainly, the correlation and information theory metrics are applied to find out the relevance criterion of a feature [39-40]. Based on the order of relevance, a rank is assigned to every feature during selection. On the contrary, the wrapper method uses a classifier to predict the significance regarding its classification correctness. However, its inherent drawback is the feature selection model may not work on other datasets using some different classifiers. Moreover, the wrapper method tends to lead overfitting issue, i.e., it may not work well on the test feature sets. The wrapper method is also computationally costly compared to the filter method. Thus, the target is to minimize the error rates while tested on the actual data. The embedded process follows similitude operation of wrapper method while the hybrid approach applies the filter and wrapper methods both. The feature selection method has been formulated next.

Consider $\mathcal{F}$ is an *n*-dimensional feature vector where the $i^{th}$ feature is denoted by $f_i \in \mathcal{F}$. According to the uniqueness, a $f_i$ is classified namely as the relevant ($f_{rel}$), irrelevant ($f_{irl}$), and redundant ($f_{red}$) [23]. The relevance of $f_i$ is determined in the training phase using a predictive distance metric such as the Euclidean. After the selection process, a subset of selected features, denoted by S=$\{f_i\}_{i=1}^{p}$ is obtained where $p<n$. The principal objective of a feature selection algorithm is to determine a right combination of relevant features from the input set $\mathcal{F}$ to attain the minimum classification error. Thus, a subset S with the minimum cardinality must be formulated that maximizes the classification accuracy ($\psi$). Now, from the perspective of forward-selection and backward-elimination related terminologies are defined.

*Relevant and irrelevant feature:* a feature $f_i$ is said to be relevant if its inclusion in the subset improves the classification accuracy during the additive forward-selection, and if removal of $f_i$ degrades the accuracy during backward-elimination. Otherwise, $f_i$ is said to be irrelevant.

*Redundant feature:* a feature $f_i$ is said to be unnecessary if its inclusion or elimination does not affect the classification accuracy significantly. It means that a $f_i$ that does not improve (while included) or reduce (while discarded) the evaluation criterion is known to be a redundant feature. Sometimes, removal of an unnecessary feature can enhance the accuracy.

From the perspective of forward-selection, relevance can be determined as

$$f_i \rightarrow \begin{cases} f_{rdt} & \text{if } \psi(S \cup \{f_i\}) - \psi(S) \approx \varepsilon \\ f_{rel} & \text{if } \psi(S \cup \{f_i\}) - \psi(S) \geq \Delta \\ f_{irl} & \text{otherwise} \end{cases} \tag{7}$$

Alternatively, from the standpoint of backward-elimination, significance can be computed as

$$f_i \rightarrow \begin{cases} f_{rdt} & \text{if } \psi(S \setminus \{f_i\}) - \psi(S) \approx \varepsilon \\ f_{rel} & \text{if } \psi(S \setminus \{f_i\}) - \psi(S) \leq \Delta \\ f_{irl} & \text{otherwise} \end{cases} \tag{8}$$

where, $\varepsilon$ and $\Delta$ represent the permissible values to determine the redundancy and relevance criterion, respectively. Generally, $\Delta > \varepsilon$ and $\varepsilon \approx 0$ is considered. It (i.e., $\Delta > \varepsilon$ ) implies that the relevance criterion also validates the checking for redundancy condition. Alternatively, instead of finding two threshold values ( $\varepsilon$ and $\Delta$ ), a single limit $\Delta$ can be used with a non-zero value that maintains $0 < \Delta < 1$.

Again, according to the present context of four-finger biometrics, a feature can also be classified into two types, namely the global, and local. The aim is to eliminate a redundant feature that reduces the cardinality of subset S by considering these two types of features. However, the researchers may have a different opinion about the classification of the global and local features in the context of hand biometrics, here, the definitions are given below.

*Global feature:* a feature is defined as global when it is attributed with the characteristic computed from the entire hand or from each finger. Example, the area of a normalized hand can be considered as a universal feature. Also, the length of every finger collectively can be termed as global feature because the same feature is attributed with the same characteristic for the four fingers at the finger-level. Thus, a single feature at the hand-level or a set of the same features one per finger at the finger-level is termed a global feature.

*Local feature:* a feature is considered as local when it is valid for only one or more particular finger(s), rather than all of the fingers. In this case, every feature is considered independently instead of a group of features. Example, the area of a particular finger is a local to that finger.

In the case of finger biometrics where only the isolated fingers are considered, a global feature is defined as

$$f_i^{gbl} \rightarrow \{f_i^{Index}, f_i^{Middle}, f_i^{Ring}, f_i^{Little}\} \tag{9}$$

In the present context, regarding the cardinality, a particular feature can also be defined as the global or local while the defined four fingers are considered.

$$f_i \rightarrow \begin{cases} f_i^{gbl} & \text{if } |f_i| = 4 \\ f_i^{lcl} & \text{if } |f_i| \leq 3 \end{cases} \tag{10}$$

A global feature ($f_{gbl}$) is called relevant during forward-selection if

$$\begin{aligned} &\psi(S \cup \{f_i^{gbl}\}) - \psi(S) \geq \Delta \\ &= \psi(S \cup \{f_i^{Index}, f_i^{Middle}, f_i^{Ring}, f_i^{Little}\}) - \psi(S) \geq \Delta \end{aligned} \tag{11}$$

Example: Let, $\mathcal{F}$ is an input set of twelve features ($|\mathcal{F}|$=12), representing the major-axis length (*ma*), area (*ar*), and solidity (*sl*) of each finger. Its representation at local to independent finger-level is given as

$$\mathcal{F} = \{f_{ma}^{Index}, f_{ma}^{Middle}, f_{ma}^{Ring}, f_{ma}^{Little}, f_{ar}^{Index}, ...., f_{ar}^{Little}, .f_{sl}^{Index} ..., f_{sl}^{Little}\} \tag{12}$$

According to the order, from the index- to little-finger, numeric labeling is followed as

$$\begin{aligned} \mathcal{F} &= \{f_{ma}^1, f_{ma}^2, f_{ma}^3, f_{ma}^4, f_{ar}^5, ...., f_{ar}^8, f_{sl}^9 .... f_{sl}^{12}\} \\ &= \{\{f_{ma}^{1-4}\}, \{f_{ar}^{5-8}\}, \{f_{sl}^{9-12}\}\} \\ &= \{\{f_{ma}^{gbl_1}\}, \{f_{ar}^{gbl_2}\}, \{f_{sl}^{gbl_3}\}\} \end{aligned} \tag{13}$$

The above formation of $\mathcal{F}$ in eq.(13) is delineated as the global representation of the features. The reason behind considering local features is to validate a feature explicitly to a finger-level which is relevant than a group of a similar feature.

Now, suppose the area of middle ($f_{ar}^{Middle}$) and index ($f_{ar}^{Index}$) fingers are relevant while the same feature of the other fingers are irrelevant or redundant. Hence, considering the areas of only these two fingers are more appropriate than selecting the areas of the four fingers. Therefore, these two precise local features reduce the cardinality of S by removing the same attribute of other fingers. Now, one question may also arise, why not to consider only the local features? The main reason is that the local feature extraction (e.g., SIFT descriptors) and selection requires a high computational cost for a large feature space. It may not also yield an optimal solution in every situation. On the other side, the global features are tested as a group which reduces the computational time regarding a number of required iterations by a factor of four (i.e., the number of fingers, or the cardinality of the group of features). Thus, there is a tradeoff between the global and local feature selection regarding the computational complexity and optimality in subset selection regarding the accuracy and cardinality of the subspace.

**Algorithm 1**: *forward-selection*

***Input:*** *Feature set:* $\mathcal{F}=\{f_i\}_{i=1}^{n}$

***Output:*** *Selected feature subset:* $S_{Fo}=\{f_i\}_{i=1}^{m}$

1. *Initialize: the number of selected features* $m \leftarrow 0$;
   *Permissible threshold accuracy to determine the relevant feature* $\Delta$; *and selected feature subset,* $S_{F0} \leftarrow \{\Phi\}$.
2. *select the most pertinent feature as the first feature* $f_1$ *from* $\mathcal{F}$
   $S_{Fo} \leftarrow f_1$
   $\mathcal{F} \leftarrow \mathcal{F} \backslash \{f_1\}$
   $m \leftarrow 1$
3. ***for*** $i \leftarrow 2,3,\ldots,n$
   ***if*** $\psi(S_{Fo} \cup f_i) - \psi(S_{Fo}) \geq \Delta$
   $S_{Fo} \leftarrow S_{Fo} \cup \{f_i\}$;
   $\mathcal{F} \leftarrow \mathcal{F} \backslash \{f_1\}$
   $i \leftarrow i+1$;
   $m \leftarrow m+1$;
   ***end if***
   ***end for***
4. *return subset* $S_{Fo}=\{f_i\}_{i=1}^{m}$

**Algorithm 2:** *backward-elimination*

***Input:*** *Feature set:* $S_{Fo}=\{f_i\}_{i=1}^{m}$

***Output:*** *Selected feature subset:* $S_{FoBa} = \{f_i\}_{i=1}^{p}$

1. *Initializen: selected feature subset* $S_{FoBa} \leftarrow S_{Fo}$
   *number of selected features* $p \leftarrow m$*;*
   *permissible accuracy to remove the irrelevant feature,* $\Delta$
2. ***for*** *i* $\leftarrow$ *1,2,….,m*
   ***if*** $\varphi(S_{FoBa} \setminus f_i) - \varphi(S_{FoBa}) \geq \Delta$
   $S_{FoBa} \leftarrow S_{FoBa} \setminus \{f_i\}$
   $i \leftarrow i+1;$
   $p \leftarrow p-1;$
   ***end if***
   ***end for***
3. *return subset* $S_{FoBa} = \{f_i\}_{i=1}^{p}$

---

- *The time complexity depends on the cardinality of the input set* $\mathcal{F}$*, and selected feature subset S. It can be estimated as* $\boldsymbol{O}(|\mathcal{F}|.|S|)$ *for both of the algorithms.*

In literature, several fundamental feature selection algorithms are devised. A comparative study has been provided by *Jain et al.* [41]. Here, the forward-selection and backward- elimination algorithms are described. A successive combination of both algorithms is formulated by *Zhang* to overcome the limitations of both methods, and the algorithm is named as the *adaptive forward-backward (FoBa)* algorithm [42]. In this greedy FoBa algorithm, forward-selection (*Algorithm-1*) is followed to find out an initial subset of relevant features. Subsequently, backward-elimination (*Algorithm-2*) is followed on the ensued subset to reduce further the cardinality and to improve the accuracy. Finally, an optimal subset S of selected features is formulated based on which experiments are conducted on the test data.

A variation of the FoBa algorithm is presented in [25], known as the *rank-based forward-backward (R-FoBa)* feature selection in the perspective of finger biometrics. In the R-FoBa algorithm, a rank is assigned to every global feature according to the relevance, evaluated by a *k*-NN classifier during the learning phase. According to the aptness, the most significant feature is delineated as the most significant (rank-1) feature. Based on the ranking order (i.e., from the highest to the lowest order of relevance), all the global features one at a time are tested for subset formulation. The R-FoBa is validated by the 10-fold cross validation using the *Pearson* correlation metric [22]. The *out-of-bag* classification error has also been estimated based on the selected subset by the bag of random classification trees [43-44]. In this method, the relevance is determined by a simple distance-based *k*-NN algorithm during training. Finally, testing on the real data is conducted by the weighted *k*-NN and RF classifiers.

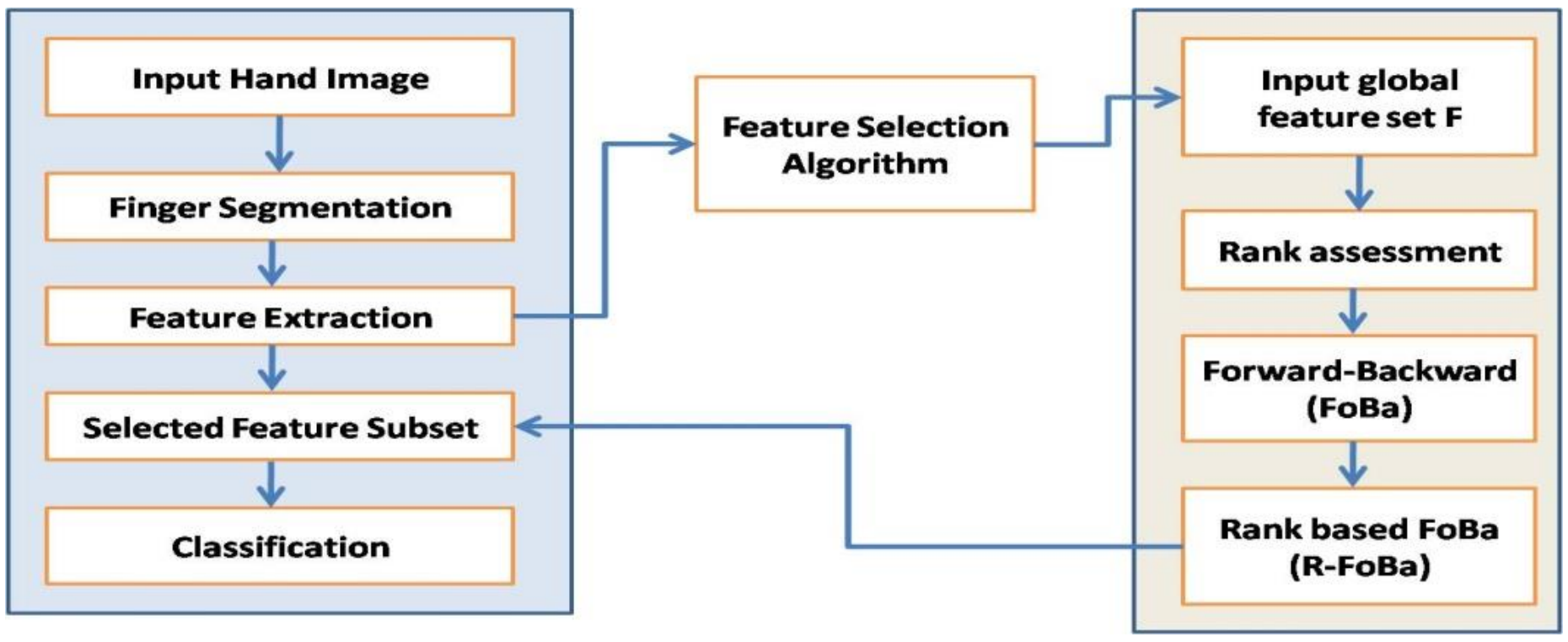

Figure 4: finger biometric system with the feature selection scheme

Another approach is the local feature selection using the same basis of FoBa, with an important target of further decrement of subset cardinality possibly with an improvement of accuracy. In this case, all the features are independently assessed for its inclusion or removal rather than a group of the same attribute. This method is named as the local feature selection. The formulation of the local subset is highlighted in the next section.

## 5 Experimentation

Firstly, a brief description of the hand database, and classifiers are presented. Next, the analyses of the various experiments are performed.

### 5.1 Database description

The Bosphorus hand database has been created for research purposes at the Boğaziçi University, Turkey [6]. This database represents the contact-free color hand images of the individuals who are the students and staff members of different universities of Turkey and France. The age variation of the subjects is between 20 and 50 years. This database contains three images per hand which has been acquired by a scanner (HP Scanjet) with 383×526 pixels at 45-dpi, at three different imaging sessions. The time lapse is between 2 weeks to 3 years, and the average time lapse is one year. The intra-class pose variation is sufficient. This publicly and freely available dataset is preferred because of its larger population space.

## 5.2 Classifier Description

The *k*-Nearest Neighbor (*k*-NN) is a widely experimented supervised algorithm, pertinent for pattern recognition and machine learning tasks. It is defined with a simple algorithmic model based on a specific distance metric with lower time complexity. The number of neighbors is scalable, and the distances between the neighborhood feature vectors can be measured using several proximity metrics such as the Euclidean, Mahalanobis, Minkowski, etc. The class-label of a query sample feature vector is determined based on the majority voting of the proximity measured from the trained data samples.

*Weighted k-Nearest Neighbor (wk-NN)*

The weight assignment to every feature is a variation of the *k*-NN algorithm to prioritize a feature over another one [45]. The weight can be determined according to the relevance of a feature using a criterion or using a suitable classification algorithm. Mainly, the distance- or correlation-based metric is followed for weight assignment. The weights are estimated based on the learning observations at the training phase and normalized within the [0, 1] range. Here, the assigned weights are based on the Euclidean distance (*d*) of a simple *k*-NN classifier with *k*=3 during the training. The weight $w_i$ of a feature $f_i$ is defined as

$$w_i = \frac{d(f_i)}{\sum_{f_i \in F} d(f_i)} \tag{14}$$

where $d(f_i)$ represents the accuracy of $f_i$ evaluated regarding the distance metric. The distance *d* between training (Tr) and testing (Ts) vector is calculated as

$$d = \sqrt{\sum_{i=1}^{n} (Tr_i - Ts_i)^2} \tag{15}$$

and $\sum_{i=1}^{n} w_i = 1$. The trained and claimed test features are denoted by $Tr_i$ and $Ts_i$, respectively. Based on the weight assignment, the distance metric is defined with the $w_i$ values as

$$d' = d.w_i = \sqrt{\sum_{i=1}^{n} (Tr_i - Ts_i)^2}.w_i \quad (16)$$

Now, the weighted *k*-NN (*wk*-NN) classifier is used for testing. Thresholding on the weight ($th_w$) is used to judge whether a feature should be included in $\mathcal{F}$ or not, i.e., $f_i \in F$ if $w_i \geq th_w$.

*Random Forest (RF)*

The random forest (RF) classification model has been introduced by *Breiman* [43-44]. The RF is based on a collection of decision trees with high precision of prediction. It is apposite for a more extensive feature space which may also represent noisy data. Every tree of the RF predicts a decision on the test dataset independently. Tree bagging method is used for classification. An unknown feature vector is tested for which a predicted score is determined based on the trained ensemble dataset. The predicted score represents the weighted average of matching probabilities determined by each classification tree of the RF. The accuracy can be improved with more ensemble trees.

## 5.3 Experimental Description

The experiments for the identification and verification are carried out for the training and testing phases which include the disjoint sets of subjects with the ratio of 2:3. During the training, 200 subjects are chosen randomly while the testing is carried out with another disjoint set of 300 subjects. Altering the subjects for training and testing with maintaining the disjoint property, several experimental sets have been formulated, and the average results are reported. Mainly, different experimental models for identification have been tested to validate the present scheme.

### *5.3.1 Identification*

The experimental scenarios are based on the feature selection methods, and the effectiveness of every finger is also assessed singly. Before conducting the significant experiments, the following initial observations are made on the actual testing set having an initial cardinality of 52 features. The *k*-NN classifier evaluates the accuracies in this method, given in Table 4. It is evident that feature normalization and weight assignment are more significant to conduct the experiments rather than using the actual values of the features.

Table 4: Identification accuracies (%) of the testing subjects

| Classifier | Subjects | actual feature set | feature normalization | weight assignment |
|---|---|---|---|---|
| *k*-NN | 300 | 93 | 94.67 | 95.67 |

Table 5: Finger-level identification accuracies (%) with weight assignment

| Phase | Subjects | Classifier | Index-finger | Middle-finger | Ring-finger | Little-finger |
|---|---|---|---|---|---|---|
| Training | 200 | *wk*-NN | 58.5 | 61 | 67.5 | 60.5 |
| | | RF | 75.5 | 76 | 82 | 81 |
| Testing | 300 | *wk*-NN | 53 | 64.34 | 65.67 | 58 |
| | | RF | 68.67 | 80.34 | 84 | 69 |

During the training, the weight is computed for each feature, and assigned accordingly. An average value is computed from the weights assigned to all features. The average value is regarded as a threshold ($th_w$=3.98) that is used to eliminate unnecessary features. If the accuracy of an independent feature is more significant than the threshold, then, the feature is chosen, otherwise delineated as an unnecessary one. This method formulates a subset of essential features representing only 19 features, rendering 91% identification accuracy by the *k*-NN classifier. However, this simple weighted-threshold is not regarded as an appropriate method for feature selection.

Next, the effectiveness of every finger (excluding the thumb) before feature selection is evaluated by the *wk*-NN and RF classifiers separately, given in Table 5. A rank is assigned to every feature during the training to illustrate the rank-based feature selection, i.e., R-FoBa presented in [25], according to the highest to the lowest order of classification accuracy rendered by the

*wk*-NN classification scheme, presented in Table 6.

Table 6: Weighted feature-level assessment for rank assignment during training

| Assessment | $f_1$ | $f_2$ | $f_3$ | $f_4$ | $f_5$ | $f_6$ | $f_7$ | $f_8$ | $f_9$ | $f_{10}$ | $f_{11}$ | $f_{12}$ | $f_{13}$ |
|---|---|---|---|---|---|---|---|---|---|---|---|---|---|
| *wk*-NN | 44 | 47 | 22 | 22 | 19.5 | 25 | 26.5 | 33.5 | 36 | 20.5 | 29 | 21 | 24.5 |
| Rank | 2 | 1 | 10 | 9 | 13 | 7 | 6 | 4 | 3 | 12 | 5 | 11 | 8 |

Table 7: Identification accuracies (%) using feature selection algorithms during training

| Classifier | Global feature selection | | | | Local feature selection | | | |
|---|---|---|---|---|---|---|---|---|
| | Cardinality | FoBa | Cardinality | R-FoBa | Cardinality | FoBa | Cardinality | R-FoBa |
| *wk*-NN | 36 | 93.5 | 40 | 94.5 | 24 | 94.34 | 25 | 95 |

Afterward, the feature selection algorithms are applied to the input set $\mathcal{F}$. For this purpose, set $\mathcal{F}$ is normalized using the *min-max* rule and weights are assigned accordingly. The features are selected using the *wk*-NN classifier during the training while the testing is conducted with the *wk*-NN and RF classifiers. Two feature selection methods, namely the global and local subsets are formulated based on the trained subject space. Also, the forward-selection is carried out by choosing the features in two ways. Firstly, the features are arbitrarily selected and tested whether the feature should be included in the subset or discarded otherwise. This scheme of randomly feature selection is regarded here as the FoBa which does not involve any prior rank assessment to avoid the pre-computation of rank. To determine the relevant feature, $\Delta = 0.003$ and for the redundant feature elimination $\varepsilon = 0$ have been followed. During the training, the outcomes of the feature selection methods based on the 200 subjects are provided in Table 7. It is evident that the R-FoBa performs better than the FoBa method concerning to the accuracy. On the other side, the dimensionality reduction of the set $\mathcal{F}$ is remarkable. It is evident that the local selection method outperforms than the global selection regarding the dimensionality optimization, i.e., the cardinality of the optimal subset. Next, using the same strategies adopted for feature selection, the test feature set is assessed, and the results are given in Table 8.

Table 8: Identification accuracies (%) using feature selection algorithms during testing

| Classifier | Global feature selection | | | | Local feature selection | | | |
|---|---|---|---|---|---|---|---|---|
| | Cardinality | FoBa | Cardinality | R-FoBa | Cardinality | FoBa | Cardinality | R-FoBa |
| *wk*-NN | 36 | 95 | 40 | 96.34 | 24 | 96 | 25 | 97 |
| RF | | 97 | | 97.67 | | 96.67 | | 98.67 |

### 5.3.2 *Overfittting issue*

Feature *overfitting* is a crucial issue regarding feature selection. Here, the 10-fold cross validation error, and the *out-of-bag* (*oob*) error are estimated based on the selected feature subsets. The 10-fold cross validation errors estimated by the *wk*-NN using the correlation metric for the global and local R-FoBa are shown in Fig.5. The comparison between the selected feature subsets of the training and testing cases are highlighted. For the testing case, altogether 900 (300 subjects × 3 samples per subject) feature vectors are partitioned, out of which 90% samples are trained, and 10% samples are used for testing at each fold of validation test. The average validation error of the ten-folds is reported here for each neighbor which has been varied from *k*=1 to *k*=5. Regarding all the cases of the neighbors, i.e., *k*=1 to 5, the average cross-validation error for the training samples is 0.079, and for the testing samples, the error is 0.073. Hence, the feature-level correlation is more significant for the test samples in this present scenario.

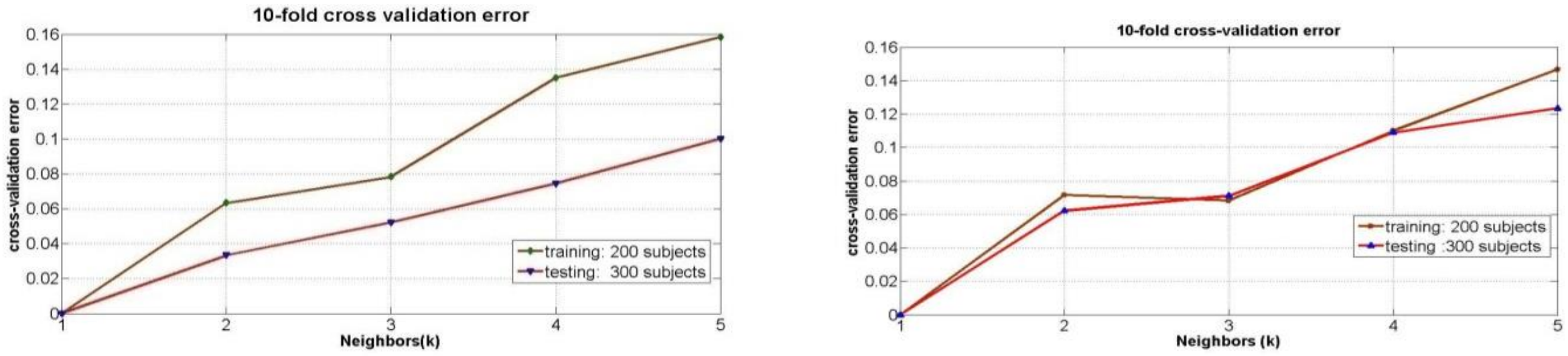

(a) (b)

Figure 5: (a) 10-fold cross validation errors by *wk*-NN using correlation. (a) global R-FoBa selection, (b) local R-FoBa selection.

Alternatively, the *oob*-errors estimated by the ensemble of random classification trees using the same local feature subsets are ideated in Fig.6. For the testing data (900 feature vectors and utmost 150 classification trees), total 900×150 observations are made. The average of mean squared error is computed for the *out-of-bag* feature vectors. In this error estimation technique, the error rates of the testing samples are lesser than the training samples too. As the number of trees grows, the *oob*-error tends to decrease usually. The average and minimum *oob*-errors are given in Table 9.

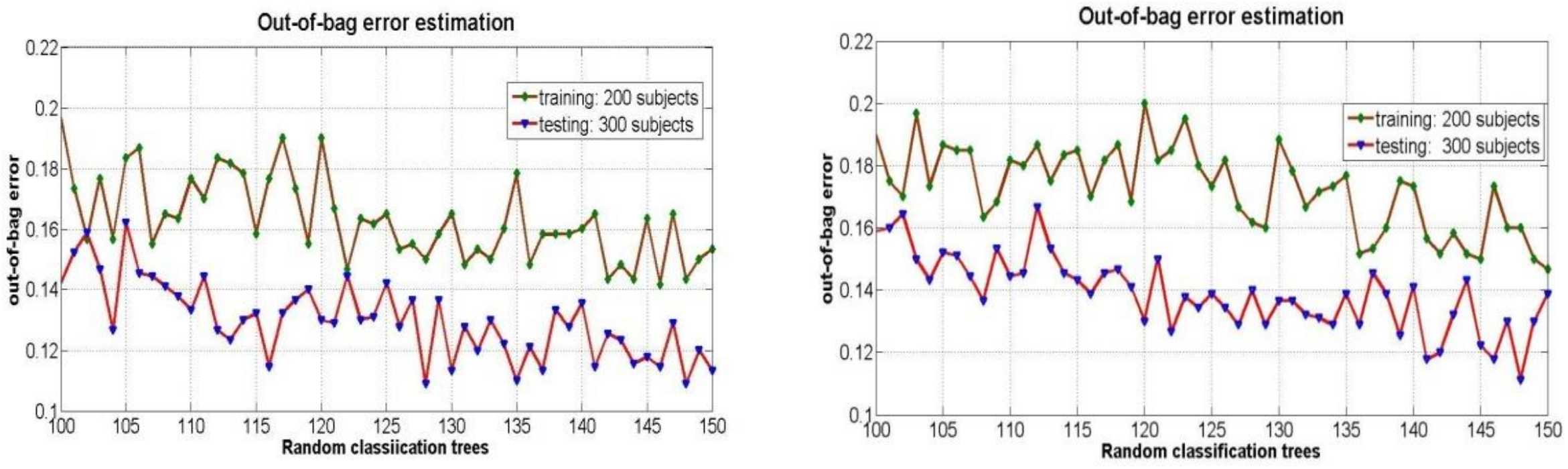

(a) (b)

Figure 6: out-of-bag error estimated by the ensemble of classification trees. (a) global R-FoBa selection, (b) local R-FoBa selection.

Table 9: Out-of-bag (oob) error estimation by the RF on the ensued subsets

| Out-of-bag error | Global R-FoBa (40 features) | | Local R-FoBa (25 features) | |
|---|---|---|---|---|
| | Average | Minimum | Average | Minimum |
| Training (200 subjects) | 0.163 | 0.141 | 0.173 | 0.146 |
| Testing (300 subjects) | 0.130 | 0.108 | 0.139 | 0.111 |

The contribution of each selected feature for both of the subsets are estimated by the *wk*-NN classifier is pictorially ideated in Fig.7. The local subset contains 25 features (marked with a vertical line in Fig.7) which finally offer 97% accuracy. Likewise, the global subset represents 40 features which collectively provide 96.34% correctness. It is evident from Fig.7 that the improvement of accuracies using the global subset, specifically after the 25$^{th}$ feature is not satisfactory. In this case, a little compromise with the accuracy can be preferred to signify the objective of dimensionality optimization. It may also reduce the computation time for the template matching during verification.

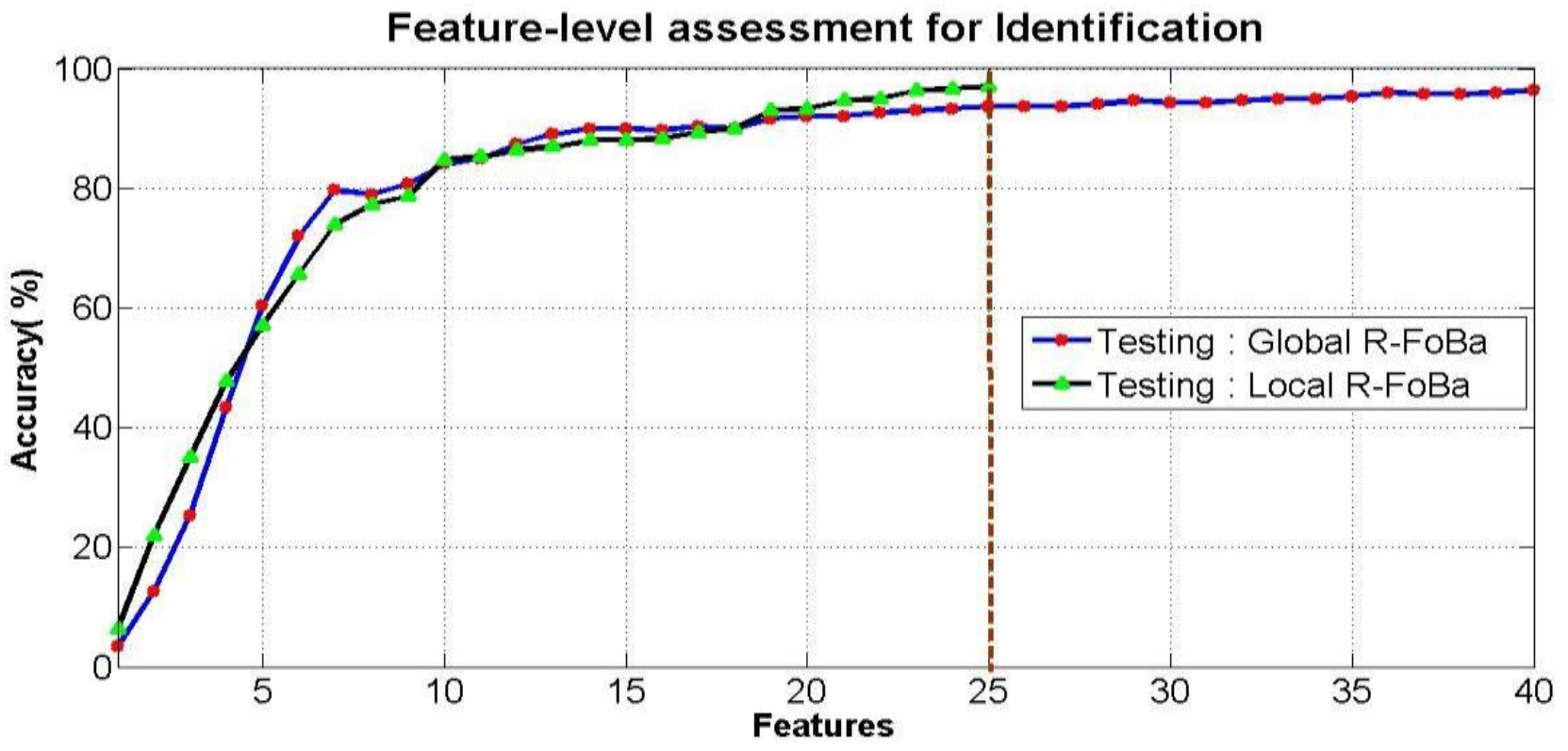

Figure 7: Feature-level accuracy assessment by the *wk*-NN classifier during testing

In another scheme, the accuracy of each finger is assessed independently. The selected features are used for finger-specific performance evaluation, and the accuracies (the number of corresponding features is mentioned in parenthesis) are provided in Table 10 and Table 11, respectively. However, the accuracies are not enhanced compared to the results provided in Table 5. The main reason is a smaller feature dimension of any particular finger. It is evident that for a little number of features, the performance cannot be improved in every situation, which is clear from the results in Table 10-11. The global selection comparatively performs better than the local selection for finger-level evaluation due to the reason of subset cardinality. However, the trade-off between the subset cardinality and accuracy is notable in this concern. The cardinality of global selection is uniform, and it is ten features per finger. Contrarily, the cardinalities of finger-level subsets for the local collection are not uniformly distributed. It can be averred that the middle-finger is the most discriminative compared to other fingers, regarding the local selection. On the other side, during the global selection the ring-finger is the most significant. Hence, either of the methods cannot be delineated as the more significant as the other. Both of the methods are optimistic based on the evaluation scheme, and the trade-off between the optimization of the feature space and accuracy.

Table 10: Global finger-level identification accuracies (%) using R-FoBa during testing

| Classifier | Index (10) | Middle (10) | Ring (10) | Little (10) |
|---|---|---|---|---|
| *wk*-NN | 45 | 62.67 | 63 | 58.67 |
| RF | 67.34 | 79 | 82.34 | 72 |

Table 11: Local finger-level identification accuracies (%) using R-FoBa during testing

| Classifier | Index (6) | Middle (8) | Ring (6) | Little (5) |
|---|---|---|---|---|
| *wk*-NN | 36.67 | 51.34 | 46 | 31.67 |
| RF | 49.34 | 70.34 | 57.34 | 46.34 |

Table 12: Verification metrics, parameters, and equal error rates during testing

| Metric | Evaluation parameter value | EER |
|---|---|---|
| Genuine Acceptance Rate (GAR): *True Acceptance / Genuine Comparison* | Training set=2×300 | Global R-FoBa: 0.066 |
| | Test set=300 | |
| False Acceptance Rate (FAR): *False Acceptance / Imposter Comparison* | Genuine comparison=600 | |
| | Imposter comparison=179400 | |
| False Reject Rate (FRR): 1- GAR | Total comparison=600×300 | Local R-FoBa: 0.046 |
| Equal Error Rate (EER): *Value at which the FAR and FRR are same.* | | |

### *5.3.3 Verification*

Verification is regarded as the one-to-one template matching scheme for performance evaluation of a biometric system. The assessment metrics regarding the verification are defined in Table 12. In the verification, a query feature template is compared to his/her stored templates on an explicit distance threshold. The distances between a claimer and the enrolled feature matrix are calculated. If the differences are within the threshold, then a person is accepted as genuine, otherwise rejected as an imposter. The threshold (T) is defined as

$$T_{e,g} = \sum_{q=1}^{n} \left(\sqrt{(\alpha_{e,q} - \beta_{g,q})^2} * w_q\right) / mean(\alpha_q) \qquad (17)$$

where, $\alpha_{e,q}$ denotes the $q^{th}$ feature of the $\alpha_e$ valid user, and $\beta_{g,q}$ means the $q^{th}$ test feature of a claimant $\beta_g$ , and $n$ represents the total number of selected features. The mean value of $q^{th}$ feature is calculated from the training dataset. The metrics and assessment parameters essential for verification have been illustrated with respect to the 300 subjects for used testing, given in Table 12. Yet again, the local R-FoBa provides better results for verification experiments regarding the EER. The receiver operating characteristic (ROC) curves are depicted in Fig.8.

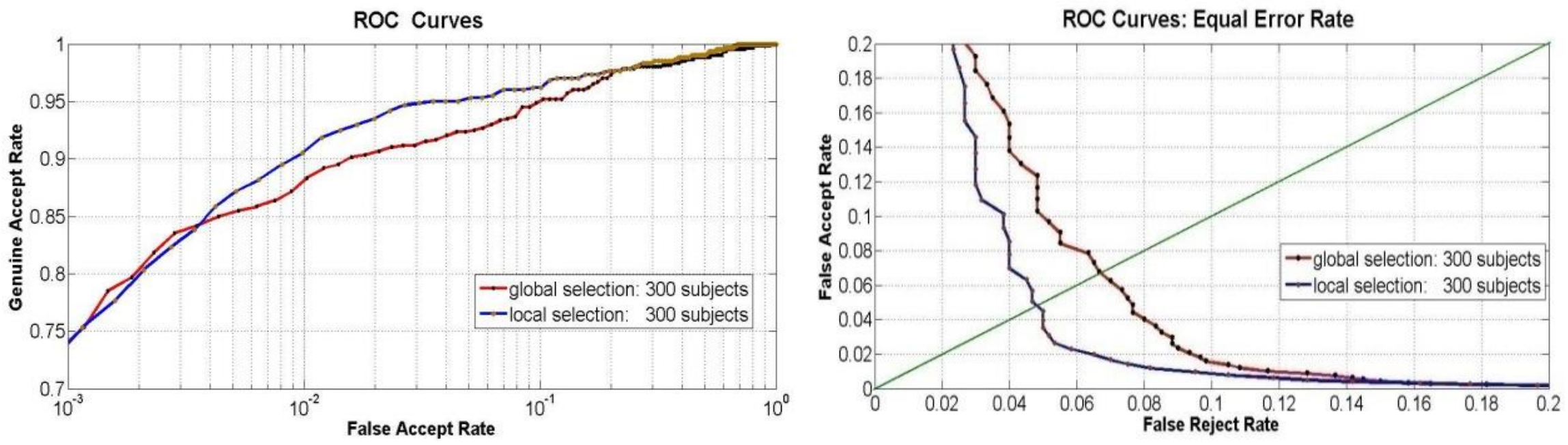

(a) (b)

Figure 8: The ROC Curves. (a) the FAR vs. GAR represented in logarithmic scale for X-axis, (b) the FRR vs. FAR to estimate the EER through the diagonal line.

In summary, the local selection performs better than the global choice. It has also been compared with the R-FoBa that the local R-FoBa provides 98.67% accuracy for the 300 subjects using the RF classifier with 25 features. Contrarily, the global R-FoBa delivers 97% accuracy with 36 features (9 selected features per finger of the left hand) in [25]. Moreover, regarding the accuracy, the present approach is also competitive with the method in [6] which provides 98.22% identification accuracy for 200 subjects with 28 geometric features. The EERs are also competitive for the cases with similar setup. Hence, from the perspective of the accuracy and subset cardinality, the local R-FoBa offers improved identification performances using the similitude experimental constraints.

## 6 Conclusion

A finger geometry recognition approach with the selected feature subset has been presented in this chapter. The prime concern is to formulate a suitable feature selection method that can render satisfactory accuracy with a smaller number of salient features. The novelty relies on the formulation of an optimal subset of features using the same notion of FoBa. The computational simplicity regarding the geometric feature extraction and selection implicates its suitability in the real world application for identity verification that mainly requires a lesser time. However, the scopes for further improvement pertinent to finger biometrics should be emphasized significantly. Primarily, the preprocessing tasks, salient feature set definition, and most importantly, selection of a right combination of a small number of features should be focused. There are also some open challenges need immediate attention regarding an accurate preprocessing phase for finger segmentation, nail-effect and clothing occlusion eradication, and others. Even, the experimental constraints can be formulated with a large number of people with sufficient posture variations. This trait can also be deployed in the mobile devices; and IoT and other online-based applications within a limited time for verification. Optimistically, it would be expected to uplift the achievements of finger biometrics by exploring these areas of advanced researches with a societal significance.